# AN OVERVIEW OF FACE LIVENESS DETECTION

Saptarshi Chakraborty[1] and Dhrubajyoti Das[2]

[1,2]Dept. of Computer Science and Engineering, National Institute of Technology, Silchar, India

## ABSTRACT

*Face recognition is a widely used biometric approach. Face recognition technology has developed rapidly in recent years and it is more direct, user friendly and convenient compared to other methods. But face recognition systems are vulnerable to spoof attacks made by non-real faces. It is an easy way to spoof face recognition systems by facial pictures such as portrait photographs. A secure system needs Liveness detection in order to guard against such spoofing. In this work, face liveness detection approaches are categorized based on the various types techniques used for liveness detection. This categorization helps understanding different spoof attacks scenarios and their relation to the developed solutions. A review of the latest works regarding face liveness detection works is presented. The main aim is to provide a simple path for the future development of novel and more secured face liveness detection approach.*

## 1. INTRODUCTION

The general public has immense need for security measures against spoof attack. Biometrics is the fastest growing segment of such security industry. Some of the familiar techniques for identification are facial recognition, fingerprint recognition, handwriting verification, hand geometry, retinal and iris scanner. Among these techniques, the one which has developed rapidly in recent years is face recognition technology and it is more direct, user friendly and convenient compared to other methods. Therefore, it has been applied to various security systems. But, in general, face recognition algorithms are not able to differentiate 'live' face from 'not live' face which is a major security issue. It is an easy way to spoof face recognition systems by facial pictures such as portrait photographs. In order to guard against such spoofing, a secure system needs liveness detection.

Biometrics is the technology of establishing the identity of an individual based on the physical or behavioural attributes of the person. The importance of biometrics in modern society has been strengthened by the need for large-scale identity management systems whose functionality depends on the accurate deduction of an individual's identity on the framework of various applications. Some examples of these applications include sharing networked computer resources, granting access to nuclear facilities, performing remote financial transactions or boarding a commercial flight [15]. The main task of a security system is the verification of an individual's identity. The primary reason for this is to prevent impostors from accessing protected resources. General techniques for security purposes are passwords or ID cards mechanisms, but these techniques of identity can easily be lost, hampered or may be stolen thereby undermine the intended security. With the help of physical and biological properties of human beings, a biometric system can offer more security for a security system.

Liveness detection has been a very active research topic in fingerprint recognition and iris recognition communities in recent years. But in face recognition, approaches are very much limited to deal with this problem. Liveness is the act of differentiating the feature space into live and non-living. Imposters will try to introduce a large number of spoofed biometrics into system. With the help of liveness detection, the performance of a biometric system will improve. It is an important and challenging issue which determines the trustworthiness of biometric system security against spoofing. In face recognition, the usual attack methods may be classified into several categories. The classification is based on what verification proof is provided to face verification system, such as a stolen photo, stolen face photos, recorded video, 3D face models with the abilities of blinking and lip moving, 3D face models with various expressions and so on. Anti-spoof problem should be well solved before face recognition systems could be widely applied in our daily life.

In the next section, a review of the most interesting face liveness detection methods is presented. Then, a discussion is presented citing the advantages and disadvantages of various face liveness detection approaches. Finally, a conclusion is drawn.

## 2. LITERATURE

There are many approaches implemented in Face Liveness Detection. In this section, some of the most interesting liveness detection methods are presented.

### 2.1 Frequency and Texture based analysis

This approach is used by Gahyun Kim et al [1]. The basic purpose is to differentiate between live face and fake face (2-D paper masks) in terms of shape and detailedness. The authors have proposed a single image-based fake face detection method based on frequency and texture analyses for differentiating live faces from 2-D paper masks. The authors have carried out power spectrum based method for the frequency analysis, which exploits both the low frequency information and the information residing in the high frequency regions. Moreover, description method based on Local Binary Pattern (LBP) has been implemented for analyzing the textures on the given facial images. They tried to exploit frequency and texture information in differentiating the live face image from 2-D paper masks. The authors suggested that the frequency information is used because of two reasons. First one is that the difference in the existence of 3-D shapes, which leads to the difference in the low frequency regions which is related to the illumination component generated by overall shape of a face. Secondly, the difference in the detail information between the live faces and the masks triggers the discrepancy in the high frequency information. The texture information is taken as the images taken from the 2-D objects (especially, the illumination components) tend to suffer from the loss of texture information compared to the images taken from the 3-D objects. For feature extraction, frequency-based feature extraction, Texture-based feature extraction and Fusion-based feature extraction are being implemented.

For extracting the frequency information, at first, the authors have transformed the facial image into the frequency domain with help of 2-D discrete Fourier transform. Then the transformed result is divided into several groups of concentric rings such that each ring represents a corresponding region in the frequency band. Finally, 1-D feature vector is acquired by combining the average energy values of all the concentric rings. For texture-based feature extraction, they used Local Binary Pattern (LBP) which is one of the most popular techniques for describing the texture information of the images. For the final one i.e. fusion-based feature extraction, the authors utilizes Support Vector Machine (SVM) classifier for learning liveness detectors with the feature vectors generated by power spectrum-based and LBP-based methods. The fusion-based method extracts a feature vector by the combination of the decision value of SVM classifier which are trained by power spectrum-based feature vectors and SVM classifier which are trained by LBP-based feature vectors. The authors have used two types of databases for their experiments: BERC Webcam Database and BERC ATM Database. All the images in webcam database were captured under three different illumination conditions and the fake faces (non-live) were captured from printed paper, magazine and caricature images. Experimental results of the proposed approach showed that LBP based method shows more promising result than frequency-based method when images are captured from prints and caricature. Overall, the fusion-based method showed best result with error rate of 4.42% compared to frequency based with 5.43% and LBP-based method with 12.46% error rate.

Similar technique of face spoofing detection from single images using micro-texture analysis was implemented by Jukka et al. [2]. The key idea is to emphasize the differences of micro texture in the feature space. The authors adopt the local binary patterns (LBP) which is a powerful texture operator, for describing the micro-textures and their spatial information. The vectors in the feature space are then given as an input to an SVM classifier which determines whether the micro-texture patterns characterize a fake image or a live person image.

The first step is to detect the face, which is then cropped and normalization is done and converted into a $64 \times 64$ pixel image. Then, they applied LBP operator on the normalized face image and the resulting LBP face image is then divided into 3×3 overlapping regions. The local 59-bin histograms obtained from each region are then computed and collected into a single 531-bin histogram. Then, two other histograms obtained from the whole face image are computed using LBP operators. Finally, a nonlinear SVM classifier with radial basis function kernel is used for determining whether the input image is a fake face or live person image. The experimental results showed that LBP has the best performance with equal error rate (EER) of 2.9% in comparison with other texture operators like Local Phase Quantization and Gabor Wavelets with EER of 4.6% and 9.5% respectively.

Another method for texture based liveness detection based on the analysis of Fourier Spectra of a single face image or face sequence image was introduced by Li et al.[13]. Their method is based on structure and movement information of live face. Their algorithm is based on two principles: first, as the size of the photo is smaller than that of live face and the photo is flat, high frequency components of photo images is less than those of real face images and secondly, even if a photo is held before a camera and is in motion, as the expressions and poses of the face contained in the photo does not vary, the standard deviation of frequency components in a sequence is small.

The authors have suggested that an effective way to live face detection is to analyze 2D Fourier spectra of the input image. They calculated the ratio of the energy of high frequency components to that of all frequency components as the corresponding high frequency descriptor (HFD). According to the authors, high frequency descriptor of the live face should be more than a reasonable threshold $T_{fd}$. The high frequency components of an image are those whose frequencies are greater than two third of the highest radius frequency of the image and whose magnitudes are also greater than a threshold $T_f$ (generally, the magnitude of high frequency components caused by the forgery process is smaller than that of original image.). The authors have found out that the above the above method will be defeated if a very clear and big size photo is used to fool the system. To solve this problem, motion images were exploited for the live face detection. So, via monitoring temporal changes of facial appearance over time, where facial appearance is represented by an energy value defined in frequency domain, is an effective approach to live face detection. The authors have proposed an algorithm which is of three steps to solve this problem. In the first step, a subset is constructed by extracting image from an input image sequence every four images. In the second step, for each image in such subset, an energy value $t$ is computed. The frequency dynamics descriptor (FDD) that is the standard deviation of the resulting flag value, is calculated for the representation of temporal changes of the face. Compared to the other works, which look for 3-D depth information of the head, the proposed algorithm has many advantages such as it is easy to compute.

Table 1. Experimental results of live face detection [13]

| Image Sequence | | Frequency Dynamics descriptor | | | High Frequency descriptor | | |
|---|---|---|---|---|---|---|---|
| | | Mean | Min | Max | Mean | Min | Max |
| Live face | 200 images | 960 | 718 | 1490 | 0.7197 | 0.4011 | 2.0544 |
| Fake face | 40 images (48x33mm) | 286 | 233 | 376 | 0 | 0 | 0 |
| | 50 images (76x55mm) | 260 | 186 | 364 | 0.0913 | 0 | 0.1376 |
| | 90 images (124x84mm) | 175 | 91 | 282 | 0.3535 | 0 | 0.5514 |
| | 20 images (600dpi) | 249 | 237 | 260 | 0.2803 | 0 | 0.3917 |

## 2.2 Variable Focusing based analysis

The technique of face liveness detection using variable focusing was implemented by Sooyeon Kim et al. [3]. The key approach is to utilize the variation of pixel values by focusing between two images sequentially taken in different focuses which is one of the camera functions. Assuming that there is no big difference in movement, the authors have tried to find the difference in focus values between real and fake faces when two sequential images(in/out focus) are collected from each subject. In case of real faces, focused regions are clear and others are blurred due to depth information. In contrast, there is little difference between images taken in different focuses from a printed copy of a face, because they are not solid. The basic constraint of this method is that it relies on the degree of Depth of Field (DoF) that determines the range of focus variations at pixels from the sequentially taken images. The DoF is the range between the nearest and farthest objects in a given focus. To increase the liveness detection performance, the authors have increased out focussing effect for which the DoF should be narrow. In this method, Sum Modified Laplacian(SML) is used for focus value measurement. The SML represents degrees of focusing in images and those values are represented as a transformed 2nd-order differential filter.

In the first step, two sequential pictures by focusing the camera on facial components are being. One is focused on a nose and the other is on ears. The nose is the closest to the camera lens, while the ears are the farthest. The depth gap between them is sufficient to express a 3D effect. In order to judge the degree of focusing, SMLs of both the pictures are being calculated. The third step is to get the difference of SMLs. For one-dimensional analysis, sum differences of SMLs (DoS) in each of columns are calculated. The authors found out that the sums of DoS of real faces show similar patterns consistently, whereas those of fake faces do not. The differences in the patterns between real and fake faces are used as features to detect face liveness. For testing, the authors have considered False Acceptance Rate (FAR) and False Rejection Rate (FRR). FAR is a rate of the numbers of fake images misclassified as real and FRR is a rate of the numbers of real images misclassified as fake. The experimental results showed that when Depth of Field (DoF) is very small, FAR is 2.86% and FRR is 0.00% but when DoF is large, the average FAR and FRR is increased. Thus the results showed that this method is crucially dependent on DoF and for better results, it is very important to make DoF small.

## 2.3 Movement of the eyes based analysis

The technique based on the analysis of movement of eyes was introduced by Hyung-Keun Jee et al. for embedded face recognition system [4]. The authors proposed a method for detecting eyes in sequential input images and then variation of each eye region is calculated and whether the input face is real or not is determined. The basic assumption is that because of blinking and uncontrolled movements of the pupils in human eyes, there should be big shape variations.

First center point of both eyes is detected in the input face image. Using detected both eyes, face region are normalized and eye regions are extracted. After binarizing extracted eye regions, each binarized eye regions are compared and variation is calculated. If the result is bigger than threshold, the input image is recognized as live face, if not, it is discriminated to the photograph. For detection of the eye regions, the authors used the fact that the intensity of the eye region is lower than the rest of face region if the image is considered as a 3D curve. To find the eye region, first, Gaussian filtering to the face image is done, so that the smoothened 3D curve is obtained. In the curve, we extract all the local minimums using the method of the gradient descent. To reduce the invalid eye candidates, the eye classifier, which is trained by Viloa's AdaBoost training methods, is used. After that, face region is being normalized by about a size and rotation by using center point of eyes because the input face can vary in size and orientation. To decrease the effect of illumination, Self Quotient Image (SQI) is applied. After Normalizing face region, eye regions are extracted based on the center of eyes. Then eye regions are binarized in order to have the pixel value of 0 and 1 by using a threshold. The threshold is obtained from the mean pixel value of each eye region. Eye regions from real faces have bigger variations in shape than regions obtained from fake faces. For calculating liveness score of each eye region, Hamming distance method is used. If two ordered lists of pixels are compared, the Hamming distance is the number of pixels that do not have same value. If the average liveness score is bigger than threshold, the input image is recognized as live face and in the case of opposite it is discriminated as a photograph.

Experimental results showed that when liveness score is measured using Hamming distance, mean score of live face is 30 and that of fake face is 17 which shows that score of live face is clearly greater than that of fake faces. So, when the threshold is set up as 21, the authors achieved best performance with FAR as 0.01 and FRR as 0.08.

Table 2. Obtained hamming distance [4]

HAMMING DISTANCE OF EYE REGIONS

|  | Hamming distance | | |
|---|---|---|---|
|  | Mean | Min | Max |
| Live face | 30 | 18 | 47 |
| Fake face | 17 | 10 | 22 |

## 2.4 Optical Flow based analysis

The method based on optical flow field was introduced by Bao et al. [5]. It analyzes the differences and properties of optical flow generated from 3D objects and 2D planes. The motion of optical flow field is a combination of four basic movement types: Translation, rotation, moving and swing.

The authors found that the first three basic types are generating quite similar optical flow fields for both 2D and for 3D images. The fourth type creates the actual differences in optical flow field. Their approach is basically based on the idea that the optical flow field for 2D objects can be represented as a projection transformation. The optical flow allows to deduce the reference field, thus allows to determine whether the test region is planar or not. For that, the difference among optical flow fields is calculated. To decide whether a face is a real face or not, this difference is being noted as a threshold. The Experiment was conducted on three groups of sample data. The first group contained 100 printed face pictures that were translated and randomly rotated, the second group contains 100 pictures from group 1 that were folded and curled before the test, the third group consisted of faces of real people (10 people, each 10 times) doing gestures like swinging, shaking, etc. The authors conducted the experiment for 10 seconds. The camera had sampling rate of 30 frames per second. The calculation was done for every 10 frames. Fig. 2 shows examples of each group ((a)-group1, (b)-group 2 and (c)-group3) as well as the results obtained.

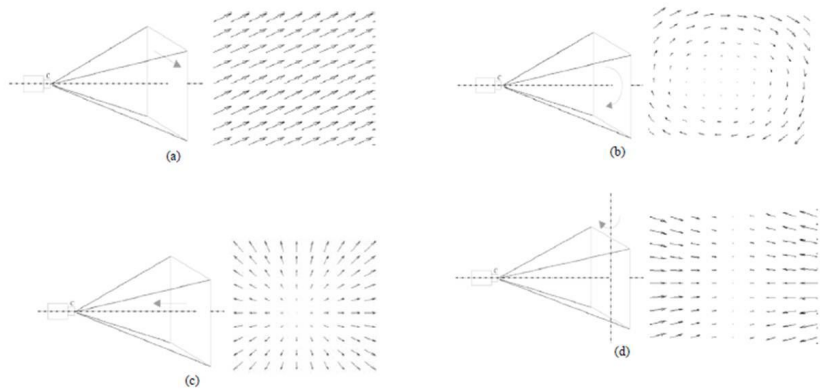

Figure 1: Four basic types of optical flow fields as presented by Bao et al. a) Translation at constant distance from the observer b) Rotation at constant distance about the view axis c) Moving forward or backward d) Swing or rotation of a planar object perpendicular to the view axis

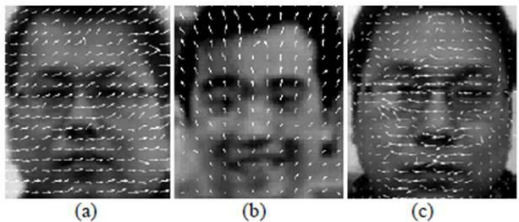

TABLE I
EXPERIMENTAL RESULTS

| Group | T | | | |
|---|---|---|---|---|
| | 0.2 | 0.4 | 0.6 | 0.8 |
| 1st | 0.54 | 0.83 | 0.86 | 0.92 |
| 2nd | 0.45 | 0.80 | 0.85 | 0.89 |
| 3rd | 1.00 | 1.00 | 0.94 | 0.86 |

Figure 2. Examples (a)-group1, (b)-group 2 and (c)-group3 and Results: the ratio of successful detection in each experiment (0 to 1)

As shown in Fig. 2, if the threshold (T) is greater, the ratio of successful detection will be higher. But at a certain point the ratio may drop, it must be noted that the authors did not mention any false acceptance rates. Another disadvantage is that illumination changes will have a negative impact on the results as the method is based on precise calculation of the optical flow field. This method will fail if the fake face is not planar i.e. it will fail for 3D objects. Therefore, authors have given advice to use this algorithm with other liveness detection methods.

A combination of face parts detection and an estimation of optical flow field for face liveness detection were introduced by Kollreider et al. [6]. This approach is able to differentiate between motion of points and motion of lines. The authors have suggested a method which analyzes the trajectories of single parts of a live face. The information which is being obtained can be used to decide whether a printed image was used or not. This approach uses a model-based Gabor decomposition and SVM for detection of face parts. The basic idea of this method is based on the assumption that a 3D face generates a 2D motion which is higher at central face regions than at the outer face regions such as ears. Therefore, parts which are farther away move differently from parts which are nearer to the camera. But, a photograph generates a constant motion on different face regions. With the information of the face parts positions and their velocity, it is possible to compare how fast they are in relation to each other [17]. This information is used to differentiate between a live face from a photograph.

The authors proposed algorithms for the computing and implementation of the optical flow of lines (OFL). For this, they have used the main Gabor filters which are linear filters for edge detection. The authors introduced two approaches for the face parts detection: first one is based on optical flow pattern matching and model-based Gabor feature classification. The second one extracts Gabor features in a non-uniform retinotopic grid and classifies them with trained SVM experts.

The database which is used contained 100 videos of Head Rotation Shot-subset (DVD002 media) of the XM2VTS database. All data were downsized to 300x240 pixels. Videos were cut (3 to 5 frames) and were used for live and non-live sequences. Each person's last frame was taken and was translated horizontally and vertically to get two non-live sequences per person. Therefore, 200 live and 200 non live sequences were examined. Most of the live sequences achieved a score of 0.75 out of 1, whereas the non-live pictures achieved a score less than 0.5. It was also noticed that glasses and moustaches lowered the score, as they were close to the camera. The authors mentioned that the system will be error free if sequences containing only horizontal movements are used. By considering a liveness score greater than 0.5 as alive, the proposed system separates 400 test sequences with error rate of 0.75 %.

Table 3. Liveness score distribution for live and non-live sequences [6]

| Liveness score | # Non-live seq. | # Live seq. |
| --- | --- | --- |
| 0 | 148 | 0 |
| 0.25 | 49 | 0 |
| 0.5 | 3 | 38 |
| 0.75 | 0 | 120 |
| 1 | 0 | 42 |

## 2.5. Blinking based analysis

The blinking-based approach for liveness detection using Conditional Random Fields (CRFs) was introduced by Lin Sun et al.[7]. The authors have used CRFs to model blinking activities, for accommodating long-range dependencies on the observation sequence. Then they compared CRF model with a discriminative model like AdaBoost and a generative model like HMM. Conditional random fields(CRFs) are probabilistic models for segmenting and labeling sequence data and mainly used in natural language processing for its accommodating long-range dependencies on the observation sequence. Blinking activity is an action represented by the image sequence which consists of images with close and non-close state.

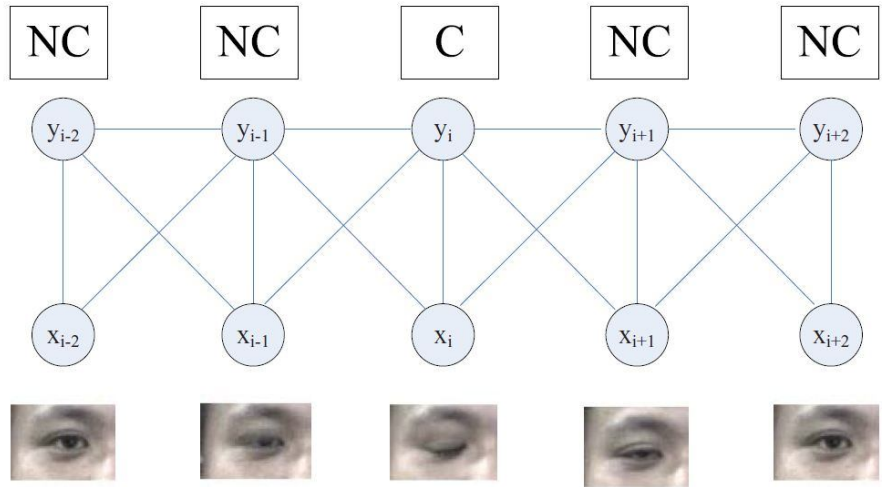

Figure 3. Graphic structure of CRF-based blinking model. Here we show the model based on contexts of observation of size 3. Labels *C* and *NC* are for close state and non-close state respectively [7]

The authors have applied a linear chain structure of CRFs. It has discrete eye state label data $y_t \in \chi = \{1, 2, \ldots, c\}$, $t = 1. . .T$, and observation $x_t$. According to the authors, Half-open state is difficult to define commonly over the different individuals, since the eye size of half-open state depends on the person eye appearance, for example, the open state of a small eye might look like the half-open state of a big eye. As for the blinking model, two state labels, *C* for close state and *NC* for non-close (include half-open and open), to label eye states are being employed by the authors. Graphical structure of CRF-based blinking model is shown in Fig. 3.

To test their approach, the authors have used video database including blinking video clips and imposter video clips, They used a total of 80 clips which is in blinking video database for 20 individuals, 4 clips for each individual: the first clip includes video without glasses in frontal view, the second clip is with thin rim glasses in frontal view, the third clip contains video with black frame glasses in frontal view, and the last clip is having video without glasses in upward view. The video clips which are being used are of five seconds' length with 30 fps and size of 320 × 240. The blinking number in each clip varies from 1 to 6 times. To test the ability against photo imposters, the authors have used 180 photo imposter video clips of 20 persons with various motions of photo, including rotating, folding and moving.

Using the blinking database, the CRF-based blinking detection is compared with cascaded Adaboost and HMM approaches. The detection rate is shown in Table 2,3,4,5.

Table 4. One-eye blinking detection rate for cascaded AdaBoost, HMM and CRF

| Different styles | Cascaded AdaBoost | HMM | CRF($w = 2$) |
|---|---|---|---|
| Without glasses | 96.5% | 73.7% | 98.2% |
| With thin rim glasses | 60.0% | 46.9% | 68.5% |
| With black frame glasses | 46.9% | 39.1% | 75.0% |
| Upward without glasses | 52.5% | 43.4% | 77.0% |

Table 5. Two-eye blinking detection rate for cascaded AdaBoost, HMM and CRF

| Different styles | Cascaded AdaBoost | HMM | CRF($w = 2$) |
|---|---|---|---|
| Without glasses | 98.2% | 82.5% | 100% |
| With thin rim glasses | 80.0% | 63.1% | 84.6% |
| With black frame glasses | 71.9% | 50.0% | 92.2% |
| Upward without glasses | 62.3% | 50.9% | 100% |

Table 6. Live face detection rate for cascaded AdaBoost, HMM and CRF

| Different styles | Cascaded AdaBoost | HMM | CRF($w = 2$) |
|---|---|---|---|
| Without glasses | 100% | 100% | 100% |
| With thin rim glasses | 95% | 95% | 90% |
| With black frame glasses | 80% | 85% | 100% |
| Upward without glasses | 85% | 95% | 100% |

Table 7. Imposter detection rate for cascaded AdaBoost, HMM and CRF

| Cascaded AdaBoost | HMM | CRF($w = 2$) |
|---|---|---|
| 95% | 97.8% | 98.3% |

Similar technique of blinking-based liveness detection was implemented by Gang Pan et al. [8]. The authors have employed the eye blink behaviors in an undirected Conditional Random Field framework, incorporated with a discriminative measure of eye states for simplifying the complex of inference and simultaneously improving the performance. The author suggested some advantages of eyeblink-based method, some of which are non-intrusion; no requirement of extra hardware .The proposed method has many advantages, one of which is that it allows relaxing the assumption of conditional independence of the observed data. Then the authors have compared their method to cascaded AdaBoost and HMM. The comparison results showed that their approach outperforms the others. However, blinking-based liveness detection has some limitations. It would be affected by strong glasses reflection, which may cover eyes partially or totally.

## 2.6 Component Dependent Descriptor based analysis

The technique of Component-based face coding approach for liveness detection was employed by Jianwei Yang et al. [9]. The authors have proposed a method which consists of four steps: (1) locating the components of face; (2) coding the low-level features respectively for all the components; (3) deriving the high-level face representation by pooling the codes with weights derived from Fisher criterion; (4) concatenating the histograms from all components into a classifier for identification.

The authors found out that significant operational difference between genuine faces and fake ones is that the former are captured by camera once, whereas the latter are obtained by re-capturing images of photos or screens. This will produce their appearance differences in three aspects: (1) Faces are blurred because of limited resolution of photos or screens and re-defocus of camera; (2) Faces appearance vary more or less for reflectance change caused by Gamma Correction of camera; (3) Face appearance also change for abnormal shading on surfaces of photos and screens. At first, the authors have expanded the detected face to obtain the holistic-face (H-Face). Then the H-Face is divided into six components (parts) which includes contour region, facial region, left eye region, right eye region, mouth region and nose region. Moreover, contour region and facial region is further divided into $2 \times 2$ grids, respectively. For all the twelve components, dense low-level features (e.g., LBP, LPQ, HOG, etc.) are extracted. Given the densely extracted local features, a component-based coding is performed based on an offline trained codebook to obtain local codes. Then the codes are concatenated into a high-level descriptor with weights derived from Fisher criterion analysis. Fisher ratio is used to describe the difference of micro textures between genuine faces and fake faces. At last, the authors feed features into a support vector machine (SVM) classifier.

For experimentation, the authors have used three different kinds of databases: NUAA Database, CASIA Database and Print-Attack Database. The authors showed that the proposed approach achieved better performance for all the databases.

## 2.7 3D Face Shape based analysis

The novel liveness detection method, based on 3D structure of the face is proposed by Andrea Lagorio et al. [10]. The proposed approach allows a biometric system to differentiate real face from a photo thus reducing the vulnerability. The authors suggested that the proposed approach can be implemented in different scenarios: either as an anti-spoofing tool, coupled with 2D face recognition systems; or can be integrated with a 3D face recognition system to perform an early detection of spoofing attacks.

The proposed algorithm computes the 3D features of the captured face data to determine if there is a live face is presented in front of the camera or not. The authors show that the lack of surface variation in the scan is one of the key evidence that the acquisition comes from 2D source. It has a very low surface curvature. Based on the computation of the mean curvature of the surface, a simple and fast method is implemented to compare the two 3D scans. An approximation of the actual curvature value at each point is computed from the principal components of the Cartesian coordinates within a given neighborhood. The mean curvature of the 3D points lying on the face surface is then computed. The authors designed two experiments. In the first one, they used FS and GVS sets. The distribution of the mean curvature values for the two sets was separated, and the value of the False Rejection Rate (FRR), was computed as zero. In the second experiment they used the FS and the Bosforus sets. In order to determine the sensitivity of the algorithm, they perform various experiments with values ranging from 4 to 20. For different values of radius, the value of the False Rejection Rate (FRR) at rank 1 is always equal to zero.

Another novel technique of face liveness detection using 3D structure recovered from a single camera was introduced by Wang et al. [20]. The authors have proposed a novel approach for face liveness detection by analyzing the sparse 3D facial structure. The basic idea is that structures from structures recovered from genuine faces usually contain sufficient 3D structure information, while structures from fake faces (photos) are usually planar in depth. From a given facial video or a sequence of images which are captured from more than two viewpoints, facial landmarks are detected and key frames are being selected. Then from the selected key frames, the sparse 3D facial structures are recovered. Then an SVM classifier is trained to differentiate live faces from fake faces. The authors have showed that the proposed approach has many advantages over many previous works. One of the advantages is that the proposed approach is independent of device and can work with various inputs. For experimentation, the authors have collected three databases using different quality cameras to inspect the anti-spoofing performance across different devices. The proposed approach achieves 100% for both classification results and face liveness detection accuracy.

## 2.8 Binary Classification based analysis

The technique of anti-spoof problem as a binary classification problem was introduced by Tan et al.[11]. The key approach which the authors have used is that a real human face is different from a face in a photo. A real face is a 3D object while a photo is 2D by itself. The surface roughness of a photo and a real face is different. The authors presented a real-time and non-intrusive method to address this based on individual images from a generic web camera. The task is being formulated as a binary classification problem, in which, however, the distribution of positive and negative are largely overlapping in the input space, and a suitable representation space is found to be of great importance. Using the Lambertian model, they proposed two strategies to extract the essential information about different surface properties of a live human face or a photograph, in terms of latent samples. Based on these, two new extensions to the sparse logistic regression model were employed which allow quick and accurate spoof detection.

For classification, the standard sparse logistic regression classifier was extended both nonlinearly and spatially to improve its generalization capability under the settings of high dimensionality and small size samples. The authors found out that the nonlinear sparse logistic regression significantly improves the anti-photo spoof performance, while the spatial extension leads to a sparse low rank bilinear logistic regression model. To evaluate their method, a publicly available large photograph-imposter database containing over 50K photo images from 15 subjects is collected by the authors. Preliminary experiments on this database show that the method proposed by the authors gives good detection performance, with advantages of realtime testing, non-intrusion and no requirement extra hardware.

Although Tan et al. have presented very effective results in their work [11]; the authors overlooked the problem of bad illumination conditions. Peixoto et al. [12] extended their work to deal with images even under bad illumination conditions either for spoof attempts coming from a laptop display or high-quality printed images. The basic key is that the brightness of the image captured from LCD screen affects the image in such a way that the high-frequency regions become prone to a "blurring" effect due to the pixels with higher values brightening their neighbourhood. This makes the fake images show less borders than the real face image.

The authors have detected whether an image is a spoof or not by exploring such information. First, they have analyzed the image using Difference of Gaussian (DoG) filter that uses two Gaussian filters with different standard deviations as limits. The basic idea of the authors was to keep the high-middle-frequencies to detect the borders in order to remove the noise. But DoG filtering does not detect the borders properly under bad illumination conditions. For the classification stage, Sparse Logistic Regression Model similar to the model in Tan et al. [11] was used by the authors. To minimize the effects of bad illumination, the image was pre-processed in order to homogenize it, so that the illumination changes become more controlled. The authors have used the contrast-limited adaptive histogram equalization (CLAHE). The main idea of CLAHE is that it operates on small regions in the image, called tiles. The Experimental results for NUAA Imposter Database of Tan et al.[11] and proposed extension for bad illumination by Peixoto et al. [12]:

Table 8. Tan et al. [11] approach

|  | Min | Mean | Max | STD |
|---|---|---|---|---|
| **Classification Accuracy** | 85.2% | 86.6% | 87.5% | 0.6% |
| **True Positive Rate** | 81.9% | 82.4% | 90.4% | 0.6% |
| **False Positive Rate** | 8.0% | 9.3% | 18.8% | 1.3% |

Table 9. Extension for bad illumination by Peixoto et al. [12]

8 tiles, Rayleigh Distribution, Contrast Limit $= 0.07$, $\alpha = 0.5$.

|                         | Min   | Mean  | Max   | STD   |
|-------------------------|-------|-------|-------|-------|
| **Classification Accuracy** | 92.0% | 93.2% | 94.5% | 0.4%  |
| **True Positive Rate**      | 92.6% | 93.0% | 93.7% | 0.4%  |
| **False Positive Rate**     | 4.7%  | 6.7%  | 8.4%  | 1.3%  |

## 2.9 Scenic Clues based analysis

The technique of face liveness detection by exploring multiple scenic clues was introduced by Yan et al. [14]. The authors have proposed a method which includes three scenic clues: non-rigid motion, face background consistency and imaging banding effect for accurate and efficient face liveness detection. Non-rigid motion clue indicates the facial motions such as blinking, and a low rank matrix decomposition based image alignment approach is implemented to extract this non-rigid motion. Face-background consistency clue assumes that the motion of face and background has high consistency for fake face images while low consistency for genuine faces, and this consistency can serve as an efficient liveness clue. The authors have implemented GMM based motion detection method for face-background consistency. Image banding effect reflects the imaging quality defects introduced in the fake face reproduction, for which the authors have used wavelet decomposition for detection. The authors have fused these three clues for efficient liveness detection.

## 2.10 Lip Movement based analysis

The liveness detection approach using lip movement classification and face detection based on face landmarks was introduced by Kollreider et al. [16]. Their work is based liveness detection using face landmarks. The proposed approach analyzes lip movements and lip reading for liveness detection. For classification of lip dynamics, SVM was used by the authors. They proposed an approach to locate the mouth regions and extract OFL in real time. They have used the XM2VTS database on various scenarios. Persons were recorded speaking digits from 0 to 9. The goal was to recognize the digits by lip-motion only. For a digit, they have used 100 short videos. For training (SVM classifier, cross validation), there were a total of 60 videos and for testing, a total of 40 videos were there. For each of digit videos, features vectors are extracted from mouth regions and given to a 10-class SVM. As a result, confusion matrices are obtained. Out of 100 individuals, recognition rate is 0.73 (73%). The authors proposed the method as an indication of liveness.

## 2.11 Context based analysis

This novel technique of context based face anti-spoofing was introduced by Komulainen et al. [18]. The authors have followed the principle of attack-specific spoofing detection and engage in face spoofing scenarios in which scene information can be exploited. They are trying to detect whether someone is trying to spoof by presenting a fake face in front of the camera in the provided view. The basic idea was that the humans rely mainly on scene and context information during the detection of spoofing; the proposed algorithm tries to impersonate human behaviour and exploits scenic cues for determining whether there a fake face is presented in front of the camera or not. The proposed approach consists of a cascade of an upper-body (UB) and a spoofing medium (SM) detector which are based on histogram of oriented gradients (HOG) descriptors and linear support vector machines (SVM). The authors suggested that the method can operate either on a single video frame or video sequences. The authors suggested an algorithm to detect close-up fake faces by describing the scenic cues with a cascade of two HOG descriptor based detectors. The alignment of the face and the upper half of the torso were examined using an upper-body detector and using a specific detector that is trained on actual face spoofing examples, the presence of the display medium is determined. To determine the proper alignment of the head-and- shoulder region, the upper-body detector that is a component of the human pose estimation pipeline is considered. For experimentation, they have used available CASIA Face Anti-Spoofing Database consisting of several fake face attacks of different natures and under varying conditions and imaging qualities. The proposed approach shows excellent performance the CASIA Face Anti-Spoofing Database showing error rate between 3.3% - 6.8%.

## 2.12 Combination of Standard Techniques based analysis

The technique that combines standard techniques in 2D face biometrics was introduced by Kollreider et al. [19]. They have looked into the matter using real-time techniques and applied them to real life spoofing scenarios in an indoor environment. First of all, the algorithm searches for faces and if the face is detected, a timer is started to define the period for collecting evidence. Then evidence is collected for the liveness detection of the faces. For liveness detection, 3D properties or eye-blinking or mouth movements in non-interactive mode are being analyzed. If no such response is found, responses are asked and checked at random. After the time period expires, verify the liveness of the face. For experimentation, a low cost web-cam that delivered 320x240 pixel frames at 25 fps was employed and computation was done on a standard laptop. The authors suggested that the performance of the proposed method is efficient for the task of public usage.

## 3 DISCUSSION

Here, liveness detection approaches are categorized based on the type of liveness indicator used to assist the liveness detection of faces. Three main types of indicators were mainly used: motion, texture and life sign.

Motion analysis mainly differentiates the motion pattern between 3D and 2D faces. It uses the fact that, planar objects move significantly different from real human faces which are 3-D objects. Motion analysis usually depends on optical flow calculated from video sequences. When using motion analysis, it is very hard to spoof by 2D face image and is independent of texture and user collaboration is not needed. But, motion analysis needs video and it is very difficult to use motion analysis when the video have low motion activity. This approach can be spoofed by 3D sculptures and it needs high quality images.

Texture analysis techniques mainly take the advantage of detectable texture patterns such as print failures, and overall image blur to detect attacks. This approach works on the assumption that fake faces are printed on paper, and the printing process and the paper structure that produce texture features can differentiate those printed images from real face images. Here, the user face is printed on a paper and presented in front of the camera for verification or identification. Using texture analysis to identify real faces is useful in such situations, as the printing procedure and paper usually contains high texture characteristics. Texture analysis based approach is easy to implement and it does not need user collaboration. But, a very diverse paper and printing textures can occur, and the systems built on texture analysis must be robust to different texture patterns which require the existence of a very diverse dataset. It is also possible that the attack is performed using a photo displayed on a screen, which will produce very low texture information.

Motion analysis will be helpful to get over the dependency on certain texture patterns. However, motion analysis may face problems when there is low motion information. This can happen because behaviour of the user may be different, high noisy images and low resolution. Motion analysis might also failed when spoof attacks is performed using more sophisticated methods, just like 3D sculpture face model.

Table 10. Advantages and Disadvantages of liveness detection approaches

| Liveness Indicator/Clue | Advantages | Disadvantages |
|---|---|---|
| Texture | 1. Easy to implement<br>2. No need of user collaboration | 1. Images with low texture information<br>2. Dataset must be diverse. |
| Motion | 1. Independent of texture<br>2. Hard to spoof by 2D image<br>3. No need of user collaboration | 1. Needs video<br>2. Difficult to use when video has low motion activity<br>3. Can be spoofed by 3D sculptures |
| Life Sign | 1. Difficult to spoof using 2D image or 3D sculptures<br>2. Independent of texture | 1. User collaboration is needed<br>2. Depends on face part detection<br>3. Needs video sequence. |

Detection of life signs can be of two types. First one assumes certain known interaction from the user. In this situation the user needs to perform a certain task to verify the liveness of his face image. This task can be a certain move that can be considered as a challenge response or a motion password. Users who will perform their task correctly are assumed to be real. The second category does not assume any interaction from the user, but focuses on certain movements of certain parts of the face, such as eye blinking, and will consider those movements as a sign of life and therefore a real face. Life sign based liveness detection based approach is very hard to spoof by 2D face images and 3D sculptures. This approach is also independent of textures but it may need user collaboration. This approach mainly depends on face part detection.

## 4 CONCLUSIONS

This work provided an overview of different approaches of face liveness detection. It presented a categorization based on the type of techniques used and types of liveness indicator/clue used for face liveness detection which helps understanding different spoof attacks scenarios and their relation to the developed solutions. A review of most interesting approaches for liveness detection was presented. The most common problems that have been observed in case of many liveness detection techniques are the effects of illumination change, effects of amplified noise on images which damages the texture information. For blinking and movement of eyes based liveness detection methods, eyes glasses which causes reflection must be considered for future development of liveness detection solutions. Furthermore, the datasets, which play an important role in the performance of liveness detection solutions, must be informative and diverse that mimics the expected application scenarios. Non-interactive video sequences must include interactive sequences where the users perform certain tasks. Future attack datasets must consider attacks like 3D sculpture faces and improved texture information. Our main aim is to give a clear pathway for future development of more secured, user friendly and efficient approaches for face liveness detection.

# REFERENCES

[1]  G. Kim, S.Eum, J. K. Suhr, D. I. Kim, K. R. Park, and J, Kim, Face liveness detection based on texture and    frequency analyses, 5th IAPR International Conference on Biometrics (ICB), New Delhi, India. pp. 67-72, March 2012.

[2]  J. Maatta, A. Hadid, M. Pietikainen, Face Spoofing Detection From Single images Using Micro-Texture Analysis, Proc. International Joint Conference on Biometrics (UCB 2011), Washington, D.C., USA.

[3]  sooyeon Kim, Sunjin Yu, Kwangtaek Kim, Yuseok Ban, Sangyoun Lee, Face liveness detection using variable focusing,  Biometrics (ICB), 2013 International Conference on, On page(s): 1 – 6, 2013.

[4]  H. K. Jee, S. U. Jung, and J. H. Yoo, Liveness detection for embedded face recognition system, International Journal of Biological and Medical Sciences, vol. 1(4), pp. 235-238, 2006.

[5]  Wei Bao, Hong Li, Nan Li, and Wei Jiang, A liveness detection method for face recognition based on optical flow field, In Image Analysis and Signal Processing, 2009, IASP 2009, International Conference on, pages 233 –236, April 2009.

[6]  K. Kollreider H. Fronthaler, and J. Bigun, Evaluating liveness by face images and the structure tensor, in Proc of 4th IEEE Workshop on Automatic Identification Advanced Technologies, Washington DC, USA, pp. 75-80, October 2005.

[7]  Lin Sun, Gang Pan, Zhaohui Wu, Shihong Lao, Blinking-Based Live Face Detection Using Conditional Random Fields, ICB 2007, Seoul, Korea, International Conference, on pages 252-260, August 27-29, 2007.

[8]  Gang Pan, Zhaohui Wu and Lin Sun, Liveness Detection for Face Recognition, Recent Advances in Face Recognition, I-Tech, on Page(s): 236, December, 2008.

[9]  Jianwei Yang, Zhen Lei, Shengcai Liao, Li, S.Z, Face Liveness Detection with Component Dependent Descriptor, Biometrics (ICB), 2013 International Conference on Page(s): 1 – 6, 2013.

[10] Andrea Lagorio, Massimo Tistarelli, Marinella Cadoni, Liveness Detection based on 3D Face Shape Analysis, Biometrics and Forensics (IWBF), 2013 International Workshop on Page(s): 1-4, 2013.

[11] X. Tan, Y. Li, J. Liu, and L. Jiang, Face liveness detection from a single image with sparse low rank bilinear discriminative model, in ECCV, 2010.

[12] B. Peixoto, C. Michelassi and A. Rocha, Face liveness detection under bad illumination conditions, In ICIP, pages 3557-3560, 2011.

[13] J. Li Y. Wang, T. Tan, and A.K. Jain, Live face detection based on the analysis of Fourier spectra, in Proc of Biometric Technology for Human Identification, Orlando, FL, USA. (SPIE 5404), pp. 296-303, April 2004.

[14] J. Yan, Z Zhang, Z. Lei, D. Yi, and S. Z. Li, Face liveness detection by exploring multiple scenic clues, In ICARCV 2012, 2012.

[15] Anil K. Jain, Patrick Flynn, Arun A. Ross, "Handbook of Biometrics", Springer, 2008.

[16] K. Kollreider, H. Fronthaler, M. I. Faraj, and J. Bigun, Real-time face detection and motion analysis with application in "liveness" assessment, IEEE Transactions on Information Forensics and Security, 2(3-2):548–558, 2007.

[17] K. Kollreider, H. Fronthaler, and J. Bigun, Non-intrusive liveness detection by face images, Image and Vision Computing, vol. 27(3), pp. 233-244, 2009.

[18] Jukka Komulainen, Abdenour Hadid, Matti Pietikainen, Context based Face Anti-Spoofing, Biometrics: Theory, Applications and Systems (BTAS), 2013 IEEE Sixth International Conference on Pages: 1-8, 2013.

[19] K. Kollreider, H. Fronthaler, and J. Bigun, Verifying Liveness by Multiple Experts in Face Biometrics, In IEEE Computer Society Conference on Computer Vision and Pattern Recognition Workshops, pages 1-6, 2008.

[20] T. Wang, J. Yang, Z. Lei, S. Liao, and S. Z. Li, Face Liveness Detection Using 3D Structure Recovered from a Single Camera, International Conference on Biometrics, Madrid, Spain, 2013.